\documentclass[sigconf]{acmart}
\renewcommand\footnotetextcopyrightpermission[1]{} 
\AtBeginDocument{%
  \providecommand\BibTeX{{%
    \normalfont B\kern-0.5em{\scshape i\kern-0.25em b}\kern-0.8em\TeX}}}

\setcopyright{acmcopyright}
\copyrightyear{2023}
\acmYear{2023}

\acmConference[Preprint]{Preprint}{2023}{Sydney}
\begin{document}

\newcommand{\ie}{\textit{i.e.}}
\newcommand{\eg}{\textit{e.g.}}

\title{Human Mobility Question Answering (Vision Paper)}

\author{Hao Xue}
\email{hao.xue1@unsw.edu.au}
\affiliation{%
  \institution{University of New South Wales}
  \city{Sydney}
  \state{NSW}
  \country{Australia}
}

\author{Flora D. Salim}
\email{flora.salim@unsw.edu.au}
\affiliation{%
  \institution{University of New South Wales}
  \city{Sydney}
  \state{NSW}
  \country{Australia}
}


\begin{abstract}
Question answering (QA) systems have attracted much attention from the artificial intelligence community as they can learn to answer questions based on the given knowledge source (\eg, images in visual question answering). However, the research into question answering systems with human mobility data remains unexplored. Mining human mobility data is crucial for various applications such as smart city planning, pandemic management, and personalised recommendation system. In this paper, we aim to tackle this gap and introduce a novel task, that is, human mobility question answering (MobQA). The aim of the task is to let the intelligent system learn from mobility data and answer related questions. This task presents a new paradigm change in mobility prediction research and further facilitates the research of human mobility recommendation systems. To better support this novel research topic, this vision paper also proposes an initial design of the dataset and a potential deep learning model framework for the introduced MobQA task. We hope that this paper will provide novel insights and open new directions in human mobility research and question answering research.
\end{abstract}




\maketitle

\section{Introduction}
In recent years, we have witnessed significant progress in the research area of human mobility data mining such as human mobility forecasting~\cite{feng2018deepmove,xue2021mobtcast}. The research of human mobility becomes an important ingredient in many smart city applications, personalised point-of-interest (POI) recommendations, etc.
In this vision paper, we are interested in shaping the next era of mobility data mining and discovering how to seamlessly bring mobility research achievements into our daily life in real-world applications for social good.
We argue that the next step for human mobility data mining is to enable a machine learning system the ability to understand human-like questions about mobility and infer corresponding answers to the questions.
We also witness that the current question answering systems are mostly limited to answering questions about vision data and text data (see Table~\ref{tab:datasets}).

Based on the above observations, in this vision paper, we propose and discuss a novel task: human mobility question answering (MobQA).
To the best of our knowledge, this is the first time that a human mobility data-based question answering task is introduced.
Note that this task is different from the existing mobility database and querying research (\eg, MobilityDB~\cite{MobilityDBTODS2020}). Although this line of work can retrieve desired information from the mobility records in the database via SQL queries, the focus has been on retrieving the exact answers using historical data stored in the database. Mobility database systems have no capability to generate future predictions beyond the given historical mobility database.
However, in MobQA, the QA system is required to synthesise information and conduct data fusion from multiple sources, involving prediction and/or recommendation tasks in addition to retrieving known answers.
The major advances to be done in MobQA research is about the mobility forecasting ability using the question answering framework.

\begin{figure}
    \centering
    \includegraphics[width=.45\textwidth]{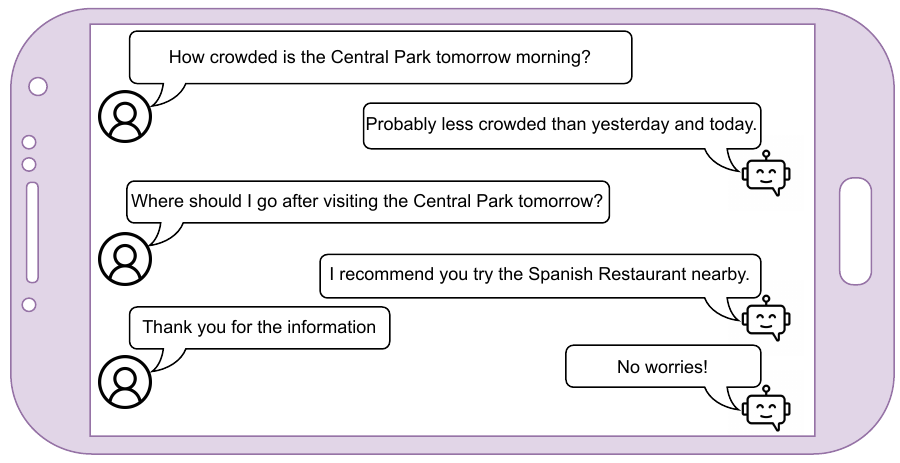}
    \caption{Concept illustration of a potential MobQA-based application: a Mobility Chatbot.}
    \label{fig:intro}
    \vspace{-3ex}
\end{figure}

\begin{table*}[]
\centering
\caption{Question Answering tasks and the corresponding knowledge sources.}
\label{tab:datasets}
\begin{tabular}{l|l} \toprule
Task/Dataset & Knowledge Source \\ \midrule\midrule
DAQUAR~\cite{malinowski2014multi}, VQA~\cite{antol2015vqa}, CLEVR~\cite{johnson2017clevr} & Image data\\
TVQA~\cite{DBLP:conf/emnlp/LeiYBB18}, ActivityNet-QA~\cite{yu2019activitynet} & Video data\\
SQuAD~\cite{DBLP:conf/acl/RajpurkarJL18}, NewsQA~\cite{DBLP:conf/rep4nlp/TrischlerWYHSBS17}, HotpotQA~\cite{DBLP:conf/emnlp/Yang0ZBCSM18} & Text data, \eg, Wikipedia paragraphs and CNN news articles \\
HybridQA~\cite{DBLP:conf/emnlp/ChenZCXWW20}, OTT-QA~\cite{DBLP:conf/iclr/ChenCSWC21} & Tabular text data, \eg, crowdsourcing-based on Wikipedia tables\\ \midrule
ForecastQA~\cite{forecastQA} & English news articles \\
TimeQA~\cite{timeqa} &  Time-evolving facts from WikiData~\cite{vrandevcic2014wikidata}\\ \midrule
\textbf{MobQA (ours)} & Human mobility data \\ \bottomrule
\end{tabular}
\vspace{-2ex}
\end{table*}

MobQA research outcomes are beneficial in multiple areas.
Firstly, the research of the MobQA task is able to spawn the next generation of deep learning algorithms for human mobility behaviour understanding. MobQA models not only can address the question answering task but also are capable to understand and forecast human mobility. 
A unified multi-task learning model, powered by the MobQA paradigm, could also bring benefits in real-world applications such as easier deployment.
Secondly, as a multi-disciplinary research task, the development of MobQA could facilitate other research directions, such as language-based mobility forecasting~\cite{xue2022translating}.
MobQA could also provide new ideas and promote novel downstream applications, such as intelligent chatbots or conversational systems with the abilities to understand mobility-related queries and also human mobility and/or spatio-temporal forecasting questions (Figure~\ref{fig:intro}). This will lead to economic and social benefits as intelligent virtual assistants are now widely embedded in many devices, environments, as well as moving vehicles.
Thirdly, the MobQA task could serve as a Turing test for mobility behaviour understanding systems. It also provides a new benchmark for evaluating existing mobility forecasting methods.

\section{Mobility Question Answering}\label{sec:mqa}

For a question answering task, there are three key components: (1) the knowledge source that is used to generate questions; (2) the questions about the knowledge; and (3) answers to the questions with the given knowledge. 
In Table~\ref{tab:datasets}, we summarise existing popular question answering tasks and their knowledge sources.
Question answering systems have attracted the attention of researchers dramatically in recent years from various domains.
As we can see from the table, the majority of existing QA tasks concentrate on: text-based question answering in the neural language processing community and image/video-based question answering in the computer vision community. 
Two QA tasks that are close to ours are the ForecastQA~\cite{forecastQA} and TimeQA~\cite{timeqa}.
The former is about exploring the forecasting of news-based events coming from general topics such as politics, sports, and economics. The latter is proposed to answer general time-related questions (\eg, the time of the second world war). Compared to these two QA tasks, MobQA focuses on the spatio-temporal mobility patterns of users visiting POIs.

Generally, our MobQA task differs from the existing QA tasks in two major aspects. First, the MobQA task is defined as a question answering system about human mobility. It aims to specifically explore how to answer questions about human mobility by using mobility data presented in the spatio-temporal format instead of commonly used text. Second, in MobQA, we are not only interested in answering questions about the given mobility records but also in answering questions about future mobility. The MobQA poses challenges in both mobility behaviour understanding and human mobility prediction.
In the following sections, we separately discuss the knowledge source (Section~\ref{sec:source}), the questions (Section~\ref{sec:q}), and the answers (Section~\ref{sec:a}) of MobQA in more detail.

\section{Dataset Design}\label{sec:dataset}
\subsection{Mobility Data Processing}\label{sec:source}
We select common publicly available human mobility datasets such as Gowalla~\cite{cho2011friendship} and FourSquare~\cite{yang2014modeling} as the raw data for the knowledge source for building MobQA dataset.
For each mobility record, we are interested in four types of information: User ID, Visited POI ID, Visited POI category, and Visited timestamp. 
In the processing of the raw dataset, we select these four columns of data. We then group the records with the same User ID and sort them by the chronological order to form this user's mobility trajectory.
Note that the User ID and the POI ID are further mapped to
random unique numbers (for the anonymisation consideration) so that no private information is included in the MobQA dataset.
For each trajectory, a sliding window is applied to split the entire trajectory into multiple clips.
Considering that we would like to incorporate prediction questions in MobQA, one trajectory clip is further divided into two segments in the temporal axis: historical observation segment $[t_{1}, t_2, \cdots, t_{\text{obs}}]$ and the ``future'' segment $[t_{\text{obs}+1}, t_{\text{obs}+2}, \cdots, t_{\text{obs}+\text{pred}}]$. Specifically, $t_{\text{obs}}$ is the observation length and $t_{\text{pred}}$ is the prediction horizon. As a result, the sliding window size equals $t_{\text{obs}} + t_{\text{pred}}$.
The historical observation segments are considered as the knowledge source data provided together with the questions as the input of the MobQA system. The future segments are mainly used to extract the ground truth answers of the prediction questions for the training/evaluation purpose 
(details given in Section~\ref{sec:a}).

\subsection{MobQA Questions}\label{sec:q}

\subsubsection{Question Formats}

Similar to other QA datasets, the MobQA dataset would consist of two formats of questions: (1) Open-ended questions: these questions cannot be simply answered with a simple ``yes'' or ``no''. This format requires the MobQA algorithm to generate more elaborate answers. (2) Multiple-choice questions: Unlike the open-ended question format that requires a free-form response, the multiple-choice task only requires an algorithm to pick from a predefined list of answer candidates.
For the open-ended format, the deep learning model is expected to output human-readable answers and this process is close to the sequence-to-sequence task setting (\eg, machine translation task). On the other hand, solving the multiple-choice questions can be considered as a typical classification task where the number of classes equals the number of answer candidates. Introducing both types of questions is valuable in facilitating the robustness of the MobQA algorithm.

\subsubsection{Question Types}

Two categories of questions should be covered in the MobQA dataset: understanding type and prediction type. Some examples of these types of questions are presented in Table~\ref{tab:template}.
For the first type, it is similar to the purpose of existing QA datasets (\eg, understanding an image in vision-based QA or understanding an article in text-based QA).
This question type is used to evaluate the algorithm's ability in understanding the given mobility behaviour data. 
The prediction type is tailored for the mobility data and the major innovation of the developed MobQA task compared to existing QA tasks. This type of question focuses on future awareness while the provided mobility data (knowledge source) describes the historical mobility observations. To successfully answer these questions, the algorithm is required to forecast future mobility based on the given historical mobility. This would be a promising function for many applications such as mobility chatbots.

\begin{table}[]
\centering
\caption{Question template examples of MobQA dataset.}
\label{tab:template}
\addtolength{\tabcolsep}{-0.75ex}
\begin{tabular}{l|p{2.25in}} \toprule
Category & Template \\ \midrule\midrule
 & Which day did the user go to \{\textit{sightseeing}\}? \\ 
Understanding & How many times did the user visit the \{\textit{Spanish Restaurant}\}? \\
 & Where did the user visit on \{\textit{Sunday morning}\}? \\ \midrule
 & Which POI is more likely going to be visited by the user on \{\textit{Wednesday noon}\}? \\
Prediction & Would the user go to \{\textit{Medical Centre}\} next \{\textit{Wednesday}\}? \\
 & Which category of POIs is more likely going to be visited by the user on \{\textit{Sunday evening}\}? \\\bottomrule
\end{tabular}
\end{table}

\subsubsection{Question Templates}
The questions in MobQA require an understanding of mobility, language, information retrieval, and commonsense knowledge to answer. Collecting interesting, diverse, unbiased, and well-posed questions is a significant challenge. 
As an alternative to the crowdsourced question collection, we argue that questions for the MobQA task can be created through question templates. Similar template-based question generation techniques have also been applied in other QA domains such as DAQUAR dataset~\cite{malinowski2014multi}.
The design of templates should match the mobility data knowledge and also be related to the mobility prediction aspect.
Table~\ref{tab:template} lists some template examples. As shown in the table, the templates include both the understanding type and the prediction type. Based on these question templates, synthetic questions can be generated automatically. According to the provided mobility data knowledge source, the entities (given in the curly brackets) could be replaced to generate a large number of questions. In addition, depending on the question formats, answer candidates can also be easily generated. Taking the last template in the table as an example, the question is asking about the POI categories. A pool that contains all the available categories can be obtained and the answer candidates are established from this pool.

\subsection{Ground Truth Answer Acquisition}\label{sec:a}
For finalising the MobQA dataset, another key step is to acquire ground truth answers which can be employed for training and evaluating the deep learning models. One advantage of the building of the MobQA dataset, different from other QA datasets, is that manually labelling and human annotators can be avoided.
Specifically, for the understanding type questions, the ground truth answer can be obtained by executing SQL queries (powered by techniques like MobilityDB~\cite{MobilityDBTODS2020}) upon the given mobility data (historical observation of the mobility records) of a particular question.
For example, for the first question template in Table~\ref{tab:template}, the ground truth answer is able to be achieved through the query: \textit{\textbf{SELECT} timestamp \textbf{FROM} observed\_mobility\_data \textbf{WHERE} category=`sightseeing'}.
Similarly, for the prediction type questions, the only difference is that the queries need to be carried out in the ``future'' mobility data segments (instead of the historical observation segments) which are also available from the raw mobility data.

\subsection{Evaluation Protocol}
For multiple-choice questions including simple yes/no questions, the evaluation process is similar to a standard classification task. Top-$k$ accuracy and F1 score should be applied. Note that the selection of $k$ depends on the size of answer candidates and top-1 accuracy is normally favourable with the 5 answer candidates case. The evaluation of open-ended questions is more complicated as multiple different answers may all be correct answers. For instance, ``shopping centre'', ``shopping mall'', or ``grocery store'' could all be considered as correct for a question. To better evaluate the closeness between the generated text and the ground truth answer corpus, we introduce the common metric for natural language generation tasks, BLEU~\cite{papineni2002bleu} as one metric for open-ended questions. In addition, the BERTScore~\cite{DBLP:conf/iclr/ZhangKWWA20}, which is computed based on the token similarity using contextual embeddings instead of exact string matches, is also considered in evaluating open-ended questions in MobQA.

\begin{figure}
    \centering
    \includegraphics[width=.4\textwidth]{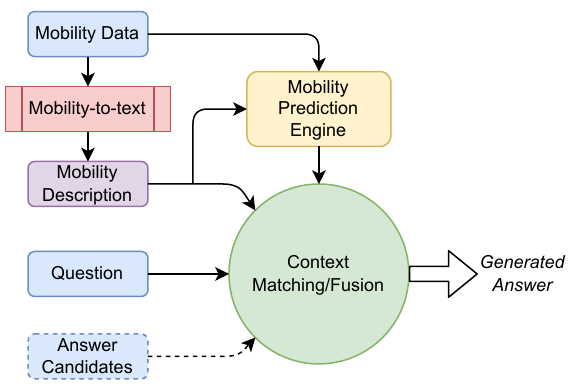}
    \caption{A potential framework design for addressing the proposed MobQA task. Note that the input of answer candidates (the dotted arrow) is not available nor used for open-ended questions.}
    \label{fig:framework}
\end{figure}

\section{Model Design for MobQA}
In this part, we introduce a visionary model design that could potentially address the MobQA task. Figure~\ref{fig:framework} shows the overview of the model. As discussed above, the input data includes three parts (shown as blue in the figure): the mobility data as the knowledge source, the question, and the answer candidates (for multiple-choice questions). Note that for open-ended questions, the answer candidates are not available.
At an initial stage, it might be difficult to simultaneously handle both the spatio-temporal mobility data given in the numerical format and the question/answer candidates given in the language format.
Hence, we hypothesise that the mobility-to-text (the red module) description developed in~\cite{xue2022translating,xue2022leveraging,xue2023promptcast} could be integrated into the MobQA model.
By applying this transformation, the knowledge source (mobility data) is converted to the context presented in the text format. As a result, the MobQA task is similar to the pure text-based open question answering task. 
Through a context matching/fusion module (the green part), the question embedding is used to query the context embedding and match answer candidates' embeddings. The understanding type questions about the mobility data can then be answered.
Since the mobility data is transferred as text, existing question answering techniques also can be leveraged for the context matching and fusion module.

Another key aspect to be addressed for the MobQA task is the ability to answer prediction-related questions. To this end, the Mobility Prediction Engine (the yellow part) is further introduced into the framework. It is the core module in the entire framework to yield future forecasting. It takes the historical mobility observations as input. Normally, the provided mobility data is considered as the input observations for the engine. From another perspective, as demonstrated in~\cite{xue2022translating}, the human mobility forecasting task can be solved in a language translation manner.
Thus, it is also a reasonable design to use the transferred mobility description in the language format or use a combination of the raw mobility data and the description as input observations like the SHIFT prediction model proposed in~\cite{xue2022translating}.
The context matching/fusion module also receives the representations from the prediction engine so that the fusion process takes the prediction into account when the framework generates answers to the prediction type questions.
By this design, the model is expected to be able to answer both understanding and prediction types of questions.

\section{Discussion and Conclusion}\label{sec:discussion}
In this paper, we present a novel human mobility question answering task which aims at better understanding mobility behaviours. Associated with this MobQA task, we discuss a blueprint on how to build a suitable MobQA dataset. Additionally, a plausible model design tailored for the MobQA task is also suggested.

\noindent\textbf{Broader Impact.}
We hope this study will offer fresh concepts and insights for human mobility research. Besides the human mobility domain, the MobQA task would also boost the corresponding research of language processing and question answering systems. For example, a new research topic could be how to develop language models suitable for sequential behaviour data. Further, the introduced MobQA task could emerge new applications such as mobility chatbots for social good.

\noindent\textbf{Future Work.}
As the first attempt to explore the question answering of human mobility data, this paper constitutes an initial step in establishing the dataset for the MobQA task. We intend to expand the question templates to include diverse questions in the dataset. After finalising the MobQA dataset, we will focus on developing deep learning models for the MobQA task based on the design introduced in this paper and evaluating our solution.

\begin{acks}
This work was supported by the Cisco Research Gift project: Mobility Question Answering (QA) for Natural Language-based Spatio-Temporal Forecasting.
\end{acks}

\bibliographystyle{ACM-Reference-Format}
\bibliography{main}

\appendix

\end{document}